\documentclass{article}

\usepackage{amsmath}
\usepackage{amsthm}
\usepackage{amssymb}
\usepackage{amsfonts}       %

\PassOptionsToPackage{numbers,compress,sort}{natbib}
\usepackage[preprint]{neurips_2024}

\usepackage[utf8]{inputenc} %
\usepackage[T1]{fontenc}    %
\usepackage[colorlinks,allcolors=black]{hyperref}       %
\usepackage{url}            %
\usepackage{booktabs}       %
\usepackage{nicefrac}       %
\usepackage{microtype}      %
\usepackage{xcolor}         %
\usepackage{bm}

\usepackage{tikz}

\usepackage{todonotes}

\usepackage[capitalize,noabbrev]{cleveref}

\theoremstyle{plain}
\newtheorem{theorem}{Theorem}[section]

\theoremstyle{definition}
\newtheorem{definition}[theorem]{Definition}
\newtheorem{assumption}[theorem]{Assumption}
\theoremstyle{remark}

\crefname{definition}{Definition}{Definitions}
\crefname{assumption}{Assumption}{Assumptions}

\DeclareFontFamily{U}{mathx}{\hyphenchar\font45}
\DeclareFontShape{U}{mathx}{m}{n}{
      <5> <6> <7> <8> <9> <10>
      <10.95> <12> <14.4> <17.28> <20.74> <24.88>
      mathx10
      }{}
\DeclareSymbolFont{mathx}{U}{mathx}{m}{n}
\DeclareMathSymbol{\bigtimes}{1}{mathx}{"91}

\title{Scaling up Probabilistic PDE Simulators with \\ Structured Volumetric Information}

\author{%
  Tim Weiland \quad Marvin Pförtner \quad Philipp Hennig \vspace{0.5em} \\
  Tübingen AI Center, University of Tübingen, Tübingen, Germany \\
  \texttt{\{tim.weiland,marvin.pfoertner,philipp.hennig\}@uni-tuebingen.de}
}

\definecolor{color1}{RGB}{0, 102, 204} %
\definecolor{color2}{RGB}{255, 102, 0}  %
\definecolor{color3}{RGB}{128, 128, 128} %

\definecolor{subdomainred}{RGB}{228, 26, 28}
\definecolor{collocationblue}{RGB}{55, 126, 184}

\begin{document}

\def\R{\mathbb{R}}
\def\N{\mathbb{N}}
\def\Rn{\mathbb{R}^n}
\def\Rd{\mathbb{R}^d}
\def\Rdout{\mathbb{R}^{d^{\prime}}}
\def\diffopD{\mathcal{D}}
\def\X{\mathbb{X}}
\newcommand{\domain}{\mathbb{D}}
\newcommand{\Lop}{\mathcal{L}}
\newcommand{\pinv}{\dagger}
\newcommand{\diag}{\textrm{diag}}
\newcommand{\tran}{\intercal}
\newcommand{\GP}{\mathcal{GP}}
\newcommand{\bigO}{\mathcal{O}}
\newcommand{\partialt}{\frac{\partial}{\partial t}}
\newcommand{\partialx}{\frac{\partial}{\partial x}}
\newcommand{\partialy}{\frac{\partial}{\partial y}}

\maketitle

\begin{abstract}
  Modeling real-world problems with partial differential equations (PDEs) is a prominent topic in scientific machine learning.
  Classic solvers for this task continue to play a central role, e.g.\ to generate training data for deep learning analogues.
  Any such numerical solution is subject to multiple sources of uncertainty, both from limited computational resources and limited data (including unknown parameters).
  Gaussian process analogues to classic PDE simulation methods have recently emerged as a framework to construct fully probabilistic estimates of all these types of uncertainty.
  So far, much of this work focused on theoretical foundations, and as such is not particularly data efficient or scalable.
  Here we propose a framework combining a discretization scheme based on the popular Finite Volume Method with complementary numerical linear algebra techniques.
  Practical experiments, including a spatiotemporal tsunami simulation, demonstrate substantially improved scaling behavior of this approach over previous collocation-based techniques.

\end{abstract}

\section{Introduction}
Partial differential equations (PDEs) are a powerful language for expressing mechanistic knowledge about the world.
Important applications include continuum mechanics \citep{Slaughter2012}, medical imaging \citep{Holder2004}, weather prediction \citep{Richardson2007}, and tsunami simulation \citep{Shuto1991}.
In the following, we focus on linear PDEs.
Let $\domain \subset \Rn$ be open and bounded. A linear PDE is expressed by an equation of the form
\begin{equation} \label{eq:def-lin-pde}
  \diffopD[u] = f,
\end{equation}
where $f: \domain \rightarrow \R$ and $\diffopD: U \subset \R^\domain \rightarrow \R^\domain$ is a linear differential operator.
We consider spatiotemporal domains in the form of $\domain := [0, T] \times \X$, $T \in \R$, and call $\X$ the spatial domain.
Additionally, we write $u(t, \bm{x})$ for functions on these domains.
Spatiotemporal physical applications impose additional constraints in the form of initial and boundary conditions.
These restrict the shape of the solution at time $t = 0$ and at the spatial domain boundaries, respectively.
A system of PDEs together with initial and boundary conditions is called an initial boundary value problem (IBVP). We seek a solution function $u \in U$ which satisfies a given IBVP.

In real-world applications, numerous sources of uncertainty affect the solution of an IBVP.
The initial and boundary conditions are often uncertain, e.g.\ due to missing data or measurement noise.
Discretizations of the PDE induce approximation error, introducing additional uncertainty.
A PDE solver that quantifies the associated uncertainties enables more informed downstream decisions.
\begin{figure*}
  \centering
  \includegraphics{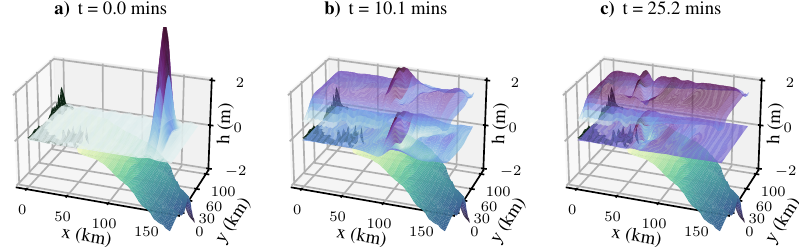}
  \caption{\textbf{A physics-informed GP models the propagation of a tsunami near a coast.} For visualization purposes, the seabed topography is \textit{not} true to scale. The middle surface (in shades of blue) depicts the posterior mean over the deviation of the water from its mean height. The transparent surface at the top depicts the $95\%$ confidence upper bound, which includes computational uncertainty caused by early termination of the internal iterative solver. We recommend looking at an animated version of this simulation \href{https://github.com/timweiland/gp-fvm}{\underline{available in our code repository.}}}
  \label{fig:tsunami}
\end{figure*}
\cref{fig:tsunami} shows an example scenario of a wave caused by an earthquake.
If we assume that sensor inaccuracies cause uncertainty about the initial displacement of the water column, this implies several different potential trajectories of a tsunami, which may be modeled by a probability distribution.

In the remainder of this work, we thus view the task of finding a solution $u$ as a machine learning problem in the sense of probabilistic numerics \citep{Hennig2015}.
Probabilistic numerical methods frame computation as Bayesian inference.
In our case, we impose a Gaussian process (GP) prior on $u$.
Then we can condition on PDE information to obtain a posterior distribution over the solution.

The theory underlying GP-based solvers for both linear and non-linear PDEs has been well-studied \citep{Cockayne2017b, Raissi2017, Chen2021, Kramer2022, Pfoertner2023}.
Various techniques have been proposed to enable inference for larger scale problems \citep{Chen2023, Yang2023}.
However, most of the existing literature focuses on point-wise observations of differential operators (which is called \textbf{collocation}).
Simulating high-frequency physical phenomena using priors based on stationary covariance functions may require small length scales.
Our work is motivated by the intuition that highly local point-wise differential observations may pose difficulties in these regimes.
Instead, we propose the use of non-local observations obtained by integrating the differential operator over a subdomain, which is equivalent to the widely used Finite Volume Method (FVM).

\textbf{Our central contribution is a framework for efficient inference on FVM observations}.
We present a class of GPs in \cref{sec:vol-inf-operators} that lends itself to particularly efficient application of FVM operators to covariance functions.
The key idea is to use box volumes and tensor product covariance functions, reducing multidimensional integrals to efficient one-dimensional integrals.
We then explore how to deal with a large number of observations in \cref{sec:large-scale-inference}, leveraging structured volume layouts to induce Kronecker structure, and combining a Cholesky decomposition with CG.
We validate the advantage of FVM observations in \cref{sec:collocation-comparison} by comparing their scaling behavior to collocation observations.
Finally, we demonstrate the real-world applicability in \cref{sec:experiments-tsunami} by computing a fully probabilistic simulation of a tsunami.

\section{Background}
\subsection{Affine Gaussian Process Inference} \label{sec:affine-inference}
The key idea underlying GP-based PDE solvers is that GPs are closed under affine transformations.
Consider transforming both functions in \eqref{eq:def-lin-pde} through a linear functional $l: U \rightarrow \R$, which yields
\begin{equation}
  l[\diffopD[u]] = l[f].
\end{equation}
If we model $u$ with a GP, then under some mild assumptions the posterior $u \mid \left( l[\diffopD[u]] = l[f] \right)$ is again a GP with closed-form mean and covariance.
In the following, we introduce the necessary notation to formalize this statement.
The underlying assumptions and theory are in Appendix \ref{appendix:affine-inference}.

\begin{definition}
  Let $\bm{\Lop_{1}}: U \rightarrow \R^{n_{1}}, \bm{\Lop_{2}}: U \rightarrow \R^{n_{2}}$ be linear operators. Define $\bm{\Lop_{1}}k: \domain \rightarrow \R^{n_{1}}$ by
  \begin{equation}
    \bm{\Lop_{1}}k(\bm{x}) := \bm{\Lop_{1}}[k(\cdot, \bm{x})]. \label{eq:Lk-def}
  \end{equation}
  Define further $\bm{\Lop_1} k \bm{\Lop_2} \in \R^{n_1 \times n_2}$ by
  \begin{equation}
    (\bm{\Lop_1} k \bm{\Lop_2^{\prime}})_{ij} := \bm{\Lop_1}[\bm{x} \mapsto \bm{\Lop_2}[k(\bm{x}, \cdot)]_j]_i. \label{eq:LkL-def}
  \end{equation}
  Similar notation may be defined for Banach space-valued linear operators \citep[Notation 2]{Pfoertner2023}.
\end{definition}
Now consider the setting where we know that the transformation of $f \sim \GP(m, k)$ under the linear operator $\bm{\Lop}: U \rightarrow \R^n$ yields observations $\bm{y} \in \R^n$ with noise $\bm{\varepsilon} \sim \mathcal{N}(\bm{\mu}, \bm{\Sigma})$.
Define $\bm{G} := \bm{\Lop} k \bm{\Lop^{\prime}} + \bm{\Sigma} \in \R^{n \times n}$ and denote by $\bm{G}^{\pinv}$ its pseudoinverse.
Then:
\begin{align}
  f \mid \bm{\Lop}[f] + \bm{\varepsilon} & = \bm{y} \sim \GP(m^{*}, k^{*}), \label{eq:affine-posterior}                                                                                   \\
  m^{*}(\bm{x})                             & := m(\bm{x}) + \bm{\Lop} k(\bm{x})^{\tran} \bm{G}^{\pinv} (\bm{y} - (\bm{\Lop} [m] + \bm{\mu})), \label{eq:lin-cond-mean}                       \\
  k^{*}(\bm{x_{1}}, \bm{x_{2}})         & := k(\bm{x_{1}}, \bm{x_{2}}) - \bm{\Lop} k(\bm{x_{1}})^{\tran} \bm{G}^{\pinv} \bm{\Lop} k(\bm{x_{2}}). \label{eq:lin-cond-cov}
\end{align}
The operator $\bm{\mathcal{I}}[u] := \bm{\Lop}[u] - \bm{y}$ is often referred to as an \textbf{information operator} \citep{Cockayne2019}.
In the following section, we show that certain PDE discretizations can be formulated through information operators. Thus, \crefrange{eq:lin-cond-mean}{eq:lin-cond-cov} enable closed-form inference from affine information.

\subsection{PDE Solving as a Learning Problem}\label{sec:mwr}
Instead of solving \cref{eq:def-lin-pde} directly, we define $n$ linear \textbf{test functionals} $l^{(i)}: U \rightarrow \R$ ($i \in \{ 1, \dots, n \}$) to obtain:
\begin{equation}
  l^{(i)}[\diffopD[u]] = l^{(i)}[f]. \quad (i \in \{1, \dots, n\}) \label{eq:mwr}
\end{equation}
In this way, an infinite-dimensional equation system is reduced to a linear equation system on $\Rn$.
This yields strictly weaker constraints on $u$, but the idea is that, as the amount of test functionals increases, the approximation error decreases.
Methods based on this approach are called \textbf{methods of weighted residuals} (MWRs), a class that includes e.g.\ finite element and Galerkin methods.

Choose test functionals $l^{(i)}$ ($i \in \{1, \dots, n\}$) and a GP prior for the solution of the PDE $u \sim \mathcal{GP}(m, k)$ such that \cref{assumption-rkbs-gaussian-randvar} is fulfilled and $\left( l^{(i)} \circ \diffopD \right)$ is a bounded linear functional on the RKBS corresponding to $u$ for all $i \in \{ 1, \dots, n \}$.
Then according to \cref{theorem:theone},
\begin{equation}
  \left(u \mid \left( l^{(i)} \circ \diffopD \right)[u] = l^{(i)}[f]\right) \sim \GP(m^{*}, k^{*}),
\end{equation}
with closed-form expressions for $m^{*}$ and $k^{*}$.
We may also combine all $n$ test functionals into one $\Rn$-valued linear operator $\bm{\Lop_{\text{PDE}}}$ instead.
The sample paths of the resulting posterior fulfill \cref{eq:mwr} for all $i \in \{ 1, \dots, n \}$.
This is a probabilistic numerical solution of the PDE.
\citet[Sec 3]{Pfoertner2023} show that this construction yields a GP with posterior mean matching MWR.

For an IBVP, the initial and boundary conditions may also be expressed through affine GP inference.
We write $\bm{\Lop_{\text{IC, BC}}}$ for the linear operator that computes all corresponding observations.
If we further write $\bm{\Lop_{\text{IBVP}}}[u] := \left( \bm{\Lop_{\text{IC, BC}}}[u] \quad \bm{\Lop_{\text{PDE}}}[u] \right)^{\tran}$, then $\bm{G} = \bm{\Lop_{\text{IBVP}}} k \bm{\Lop^{\prime}_{\text{IBVP}}} + \bm{\Sigma}$ is the Gram matrix of interest, and GP inference serves as a full probabilistic pipeline to numerically solve an IBVP.

\newcommand\redxmark[1][]{%
  \tikz\draw[thick, line cap=round,x=1ex,y=1ex,color=red,#1]
  (0,0) -- ++(45:1)
  (0,0) -- ++(-45:1)
  (0,0) -- ++(-135:1)
  (0,0) -- ++(135:1);
}

\newcommand\redinterval[1][]{%
  \tikz\draw[thick, line cap=round,x=1ex,y=1ex,color=red,#1]
  (0,0) -- (2,0)
  (0,0) -- (-2,0)
  (2,0.6) -- (2,-0.6)
  (-2,0.6) -- (-2,-0.6);
}
\section{Volumetric Information Operators} \label{sec:vol-inf-operators}
\begin{figure*}
  \centering
  \includegraphics{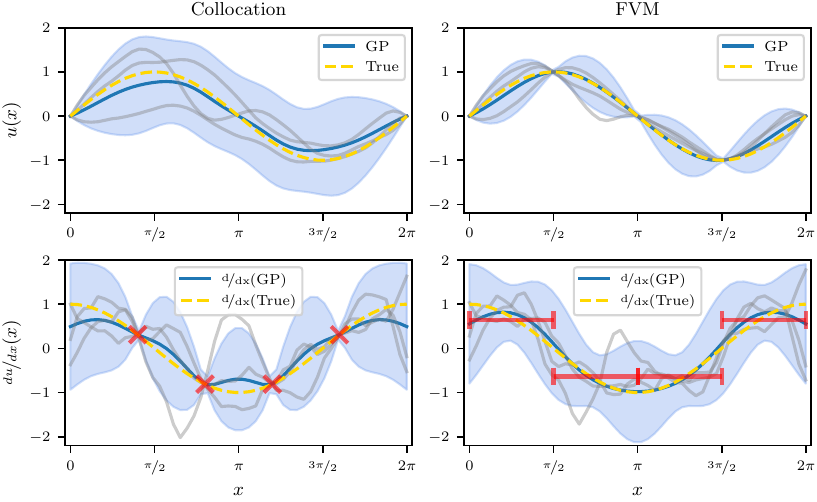}
  \caption{\textbf{Learning to solve $\nicefrac{\mathrm{d}u}{\mathrm{d}x}(x) = \cos(x)$ through collocation and FVM.} Both approaches condition a GP prior on i) the boundary conditions $u(0) = u(2 \pi) = 0$ and ii) a discretization of the differential equation. Collocation discretizes via point observations \redxmark, whereas FVM discretizes via integral observations \redinterval. The plots depict the resulting posteriors over $u(x)$ and $\nicefrac{\mathrm{d}u}{\mathrm{d}x}(x)$, as well as the corresponding ground truths. We also plot \textcolor{gray}{three samples} for each posterior.}
  \label{fig:collocation-vs-fvm}
\end{figure*}

Most GP-based PDE solvers are based on test functionals $l_{\bm{x}}[u] = u(\bm{x})$, inducing the collocation linear operator $\mathcal{L}_{\bm{x}}[u] = \diffopD[u](\bm{x})$.
In this work, we instead consider test functionals of the form $l_V[u] = \int_{V} u(\bm{x}) d\bm{x}$, where $V \subset \domain$ is some chosen volume.
Affine inference involves transforming the covariance function under $l_V \circ \diffopD$, which we call the Finite Volume Method (FVM) operator:
\begin{equation}
  \Lop_{V}[u] := \int_{V} \diffopD[u](\bm{x}) d\bm{x}. \label{eq:linop-def}
\end{equation}
Mild assumptions on $u \sim \GP(m, k)$ yield:
\begin{equation}
  \left(u \mid \int_{V} \diffopD[u](\bm{x}) d\bm{x} = \int_V f(\bm{x})d\bm{x} \right) \sim \GP(m^{*}, k^{*}),
\end{equation}
with closed-form expressions for $m^{*}$ and $k^{*}$ (\crefrange{eq:lin-cond-mean}{eq:lin-cond-cov}).
The samples of the resulting posterior fulfill the PDE on average across the chosen volume $V$.
The method resulting from this choice of test functional is classically called the subdomain method.
As stated in \citet{Fletcher1984}, the subdomain method is equivalent to the Finite Volume Method (FVM) due to the divergence theorem. %
Hence, we use the name \textbf{GP-FVM} for GPs conditioned on FVM observations in the remainder of this work.
\cref{fig:collocation-vs-fvm} shows the difference between collocation and GP-FVM on an example problem.

GP-FVM requires transforming the covariance function under $\Lop_V$.
In the following, we explore a framework to enable efficient inference from FVM observations.
We start with functions $u: \domain \subset \R \rightarrow \R$ and build up towards functions $\bm{u}: \domain \subset \Rd \rightarrow \Rdout$.
Appendix \ref{appendix:theoretical-justification} justifies that the constructions of this section satisfy the assumptions underlying \crefrange{eq:affine-posterior}{eq:lin-cond-cov}.

\subsection{One-dimensional domains}
\label{sec:one-dim}

For $\domain \subset \R$, we consider intervals of the form $V := [ a, b ]$.
Thus, the FVM operator has the form
\begin{equation}
  \mathcal{L}_{[a, b]}[u] = \int_{a}^{b} \diffopD[u](x) dx. \label{eq:linop-def-univariate}
\end{equation}
A linear differential operator $\diffopD$ is a linear combination of partial derivatives up to order $k \in \N$:
\begin{equation}
  \diffopD[u] = \sum_{i=0}^{k} c_{i} \frac{d^{i}u}{dx^{i}}.
\end{equation}
Hence, \cref{eq:linop-def-univariate} is equivalent to
\begin{equation}
  \mathcal{L}_{[a, b]}[u] = \sum_{i=0}^{k} c_{i} \int_{a}^{b} \frac{d^{i}}{dx^{i}} u(x) dx
  =  c_{0} \int_{a}^{b} u(x) dx + \sum_{i=0}^{k-1} c_{i+1} \left( u^{(i)}(b) - u^{(i)}(a) \right). \label{eq:linop-simplified}
\end{equation}
The fundamental theorem of calculus simplifies the terms where integrals are mixed with derivatives to differences of derivatives.
GP-FVM thus requires efficient access to the integral (for $i = 0$) and arbitrary derivatives $u^{(i)}$ of the covariance function.
The Matérn covariance function is widely used in the context of GP regression.
It is an isotropic covariance function of the form $k_{\nu, \ell}(x_1, x_2) = k_{\nu}(\lvert x_1 - x_2 \rvert / \ell)$ ($\nu \in \R_+$).
For $\nu = q + \frac{1}{2}$ ($q \in \N$), it simplifies to
\begin{equation}
  k_{\nu = q + \frac{1}{2}}(r) = \exp \left( - \sqrt{2q+1} r \right) p_{q} \left( \sqrt{2q+1} r \right),
\end{equation}
where $p_{q}$ is a polynomial of order $q$ with known coefficients \citep{Rasmussen2005}.
Application of the product rule shows that derivatives again have the form of an exponential times a polynomial.
Thus, evaluations of the Matérn covariance function and arbitrary derivatives thereof are efficient.
One can also derive efficient closed-form integrals of the Matérn covariance function.

\subsection{Multi-dimensional domains}
\label{sec:multi-dim}
We would like to transfer the efficiency of the one-dimensional case to a multi-dimensional input domain $\domain \subset \Rd$.
The idea is to reduce multi-dimensional integrals to one-dimensional integrals.
To achieve this, we restrict ourselves to box volumes of the form $V := [a_{1}, b_{1}] \times \dots \times [a_{d}, b_{d}]$.
More complex shapes can be approximated by unions of box volumes.
Due to Fubini's theorem, \cref{eq:linop-def} can then be expressed in terms of one-dimensional integrals.
Furthermore, we write the order-$k$ linear differential operator $\diffopD$ as a linear combination of partial derivatives. This yields:
\begin{equation}
  \Lop_{V}[u] = \int_{a_{1}}^{b_{1}} \dots \int_{a_d}^{b_{d}} \diffopD u(\bm{x}) dx_d \dots dx_1 = \sum_{\lvert \bm{\alpha} \rvert \leq k} c_{\bm{\alpha}} \int_{a_{1}}^{b_{1}} \dots \int_{a_{d}}^{b_{d}} D^{\bm{\alpha}}u(\bm{x}) dx_{1} \dots dx_{d}.
\end{equation}

To harness this structure, we use a tensor product of one-dimensional Matérn covariance functions:
\begin{equation}
  \label{eq:tensorproduct-cov-def}
  k(\mathbf{x_{1}}, \mathbf{x_{2}}) := \prod_{i=1}^{d} k^{(i)}_{\nu_{i}, \ell_{i}}(x_{1i}, x_{2i}).
\end{equation}
This allows us to distribute the partial derivatives and the one-dimensional integrals to the individual one-dimensional covariance functions.
For two different box volumes $V, V^{\prime} \subset \domain$, we get
\begin{align}
  \label{eq:multi-input-fv}
  (\Lop_{V} k)(\mathbf{x_{2}})          & = \sum_{\lvert \bm{\alpha} \rvert \leq k} c_{\bm{\alpha}} \prod_{i=1}^{d} \Lop_{[a_{i}, b_{i}]; \alpha_i}[k^{(i)}_{\nu_{i}, l_{i}}](x_{2i}), \\
  \Lop_{V} k \Lop^{\prime}_{V^{\prime}} & = \sum_{\substack{\lvert \bm{\alpha} \rvert \leq k                                                                                      \\ \lvert \bm{\alpha^{\prime}} \rvert \leq k^{\prime}}} c_{\bm{\alpha}} c_{\bm{\alpha^{\prime}}} \prod_{i=1}^{d} \Lop_{[a_{i}, b_{i}]; \alpha_i} k^{(i)}_{\nu_{i}, l_{i}} \Lop^{\prime}_{[a^{\prime}_i, b^{\prime}_i]; \alpha^{\prime}_i}, \quad\text{with} \label{eq:LkL}\\
  \Lop_{[a, b]; k}[u]                   & := \int_{a}^{b} \frac{d^{k}}{d x^{k}} u(x) dx.
\end{align}
In this way, the multi-dimensional integrals reduce to linear combinations of products of one-dimensional integrals of the type introduced in \cref{sec:one-dim}.
The smoothness properties of the one-dimensional covariance functions are transferred to the tensor product:
If $\nu_{i} = \beta_{i} + \frac{1}{2}$ ($\beta_{i} \in \N_{0}$) and $m \in C^{\bm{\beta}}(\bar{\domain})$ (prior mean), then the sample paths (almost surely) lie in $C^{\bm{\beta}}(\bar{\domain})$ \citep{Wang2021, DaCosta2023}. %
Thus, this prior ensures manageable smoothness and enables efficient inference from box volume FVM observations, eliminating the requirement for costly multi-dimensional numerical quadrature.

\subsection{Multi-dimensional codomains}
So far we assumed a one-dimensional codomain. As a final step, we extend to multi-dimensional codomains to model functions $\bm{u}: \domain \subset \Rd \rightarrow \Rdout$.
A multi-output GP prior for this situation uses a covariance function $\bm{k}: \R^{d} \times \R^{d} \rightarrow \R^{d^{\prime} \times d^{\prime}}$.
The off-diagonal entries of the output matrix are the cross-covariances between different outputs \citep[e.g.][]{Alvarez2012}.
To extend our framework to the multi-output case, we simply set the cross-covariances between the outputs to zero, i.e.
\begin{equation}
  \bm{k}(\bm{x_{1}}, \bm{x_{2}}) = \diag \left( k_{11}(\bm{x_1}, \bm{x_2}) \dots k_{d^{\prime} d^{\prime}}(\bm{x_1}, \bm{x_2}) \right), \label{eq:multioutput-cov-def}
\end{equation}
where the $k_{ii}: \R^{d} \times \R^{d} \rightarrow \R$ are tensor product covariance functions as in \cref{eq:tensorproduct-cov-def}.
This prior assumes that the outputs are uncorrelated, which is clearly not correct for realistic PDEs with several output variables.
But, these dependencies can still be modelled in the posterior, their omission in the prior is merely for simplicity and efficiency \citep[for alternatives, see][]{Alvarez2012}.
A multi-dimensional codomain yields functions $\bm{u}(\bm{x}) = \left( u_1(\bm{x}) \dots  u_{d^{\prime}}(\bm{x})\right)^{\tran}$.
In this case, linear differential operators take the form
\begin{equation}
  \diffopD[\bm{u}] := \sum_{\lvert \bm{\alpha} \rvert \leq k} \sum_{i=1}^{d^{\prime}} c_{\bm{\alpha}, i} D^{\bm{\alpha}} [u_i].
\end{equation}
$\diffopD \bm{k}(\bm{x_1}, \bm{x_2})$ is then \citep[Remark 4.1]{Pfoertner2023} a linear combination of vectors $D^{\bm{\alpha}} \bm{k_{i, :}}(\bm{x_1}, \bm{x_2}) \in \R^{d^{\prime}}$ with:
\begin{equation}
  (D^{\bm{\alpha}} \bm{k_{i, :}}(\bm{x_1}, \bm{x_2}))_j := \delta_{ij} \cdot \left(D^{\alpha}k_{jj} \right)(\bm{x_1}, \bm{x_{2}}).
\end{equation}
Similarly, $\diffopD \bm{k} \diffopD^{\prime}(\bm{x_1}, \bm{x_2})$ is a linear combination of scalars $D^{\bm{\alpha_1}} k_{i_1 i_2} {D^{\bm{\alpha_2}}}^{\prime}(\bm{x_1}, \bm{x_2})$ with:
\begin{equation}
  D^{\bm{\alpha_1}} k_{i_1 i_2} {D^{\bm{\alpha_2}}}^{\prime}(\bm{x_1}, \bm{x_2}) := \delta_{i_1 i_2} \cdot D^{\bm{\alpha_1}} k_{i_1 i_1} D^{\bm{\alpha_2}}(\bm{x_1}, \bm{x_2}).
\end{equation}
Hence, there is no added complexity over the computations from \cref{sec:multi-dim}.

\section{Large-Scale Inference}
\label{sec:large-scale-inference}
Conditioning on FVM observations in the framework described in \cref{sec:vol-inf-operators} produces a GP posterior with closed-form expressions for the posterior mean and covariance given by \crefrange{eq:lin-cond-mean}{eq:lin-cond-cov}.
The challenge considered in this section is the computation of the quantities involved in these expressions.
The computation of the posterior mean (\cref{eq:lin-cond-mean}) requires one linear system solve involving the Gram matrix $\bm{G}  \in \R^{N \times N}$ to obtain the \textbf{representer weights}:
\begin{equation}
  \bm{\alpha} := \bm{G}^{\pinv} (\bm{y} - (\bm{\Lop}[m] + \bm{\mu})).
\end{equation}
Evaluating the posterior covariance (\cref{eq:lin-cond-cov}) at $N^{*}$ test points requires $N^{*}$ solves with $\bm{G}$.
The classic approach to solve these linear systems is to compute the Cholesky decomposition $\bm{G} = \bm{L} \bm{L}^{\tran}$, where $\bm{L} \in \R^{N \times N}$ is lower triangular with positive diagonal entries.
This requires $\frac{1}{3} N^3$ FLOPs to leading order.
Afterwards, linear system solves involving $\bm{G}$ require $\bigO(N^2)$ operations using forward and backward substitution.
However, for the use case of PDE solving, naive computation of the Cholesky decomposition quickly becomes infeasible. An example resolution of $100^3$ finite volumes for a 3D domain already induces Gram matrices with $N \geq 10^6$. The corresponding Cholesky decomposition requires terabytes of RAM and a leading order of $\frac{1}{3} \cdot 10^{18}$ FLOPs to compute.

\subsection{Iterative techniques}
Iterative methods such as the conjugate gradients (CG) method address this problem.
CG approximates a solution to $\bm{Gx} = \bm{b}$ by iteratively searching in $\bm{G}$-orthogonal directions formed from the gradient of the objective function $\bm{x} \mapsto \frac{1}{2} \bm{x}^{\tran} \bm{G} \bm{x} - \bm{x}^{\tran} \bm{b}$.
CG can be implemented as a matrix-free method as it only requires access to matrix-vector products with $\bm{G}$.
We use an iterative method called IterGP \citep{Wenger2022}.
IterGP generalizes CG (among other methods) and produces a combined posterior.
The combined posterior adds \textbf{computational uncertainty} to the mathematical uncertainty from \cref{eq:lin-cond-cov}.
As CG obtains a better estimate of the representer weights, the computational uncertainty decreases.
Thus, IterGP produces an \textbf{exact} posterior even after $k \ll N$ iterations by quantifying the uncertainty induced by early termination through combined posterior uncertainty.

Iterative methods are especially effective when fast matrix-vector products with $\bm{G}$ are available.
To achieve this, we use structured discretization schemes.
\begin{definition}
  A box FVM discretization scheme is a non-empty set $\mathcal{V}$ of box volumes.
  We say that $\mathcal{V}$ has factorized structure if
  \begin{equation}
    \mathcal{V} = \{ I_1 \times \dots \times I_d  \mid I_i \in G_i \text{ for } i = 1, \dotsc, d \},
  \end{equation}
  where with $i \in \{1, \dots, d\}$ and $N_i \in \N$: $\mathcal{G}_i = \{ [a_{i, 1}, b_{i, 1}], \dots, [a_{i, N_i}, b_{i, N_i}] \}$.
\end{definition}
Define ``stacked'' linear operators $\bm{\Lop_{\mathcal{V}}}, \bm{\Lop_{\mathcal{G}_i}}$ through $\bm{\Lop_{\mathcal{V}}}[u] \in \R^{\lvert \mathcal{V} \rvert}, \left(  \bm{\Lop_{\mathcal{V}}}[u] \right)_j = \Lop_{V_j}[u]$ and $\bm{\Lop_{\mathcal{G}_i}}[u] \in \R^{N_i}, \left(  \bm{\Lop_{\mathcal{G}_i}}[u] \right)_j = \Lop_{[a_{i, j}, b_{i, j}]}[u]$.
Then $\bm{\Lop_{\mathcal{V}}} k \bm{\Lop^{\prime}_{\mathcal{V}}} \in \R^{\lvert \mathcal{V} \rvert \times \lvert \mathcal{V} \rvert}$ contains the cross-covariances between the volumes. If $\mathcal{V}$ has factorized structure, \cref{eq:LkL} shows that $\bm{\Lop_{\mathcal{V}}} k \bm{\Lop^{\prime}_{\mathcal{V}}}$ has Kronecker structure:
\begin{equation}
  \label{eq:kronecker-structure}
  \bm{\Lop_{\mathcal{V}}} k \bm{\Lop^{\prime}_{\mathcal{V}}} = \sum_{\lvert \bm{\alpha} \rvert \leq k, \lvert \bm{\alpha^{\prime}} \rvert \leq k } c_{\bm{\alpha}} c_{\bm{\alpha^{\prime}}} \bigotimes_{i=1}^{d} \bm{\Lop_{\mathcal{G}_i}} k^{(i)}_{\nu_{i}, l_{i}} \bm{\Lop^{\prime}_{\mathcal{G}_i}}.
\end{equation}
Matrix-vector products with Kronecker products are cheap and only require access to the individual factors \citep{VanLoan2000}.
For reasonable values of $\lvert \mathcal{G}_i \rvert$, we can precompute $\bm{\Lop_{\mathcal{G}_i}} k^{(i)}_{\nu_{i}, l_{i}} \bm{ \Lop^{\prime}_{\mathcal{G}_i}}$.
In this way, even discretization schemes with hundreds of thousands of volumes remain tractable.

\subsection{Cholesky preconditioners}
\label{sec:cholesky-preconditioner}
The solution of an IBVP needs to satisfy initial and boundary conditions in addition to the PDE.
Thus, the Gram matrix is constructed from different types of observations.
Applying CG directly to such an inhomogeneous matrix may require many iterations to obtain accurate representer weights.
We consider situations where a reasonable amount of observations is sufficient to cover the initial and boundary conditions, making a Cholesky decomposition feasible.
This is possible even for large-scale problems, exemplified by the experiment in \cref{sec:experiments-tsunami}.
In these cases, we can separate the initial and boundary observations from the PDE observations.
First, compute the Cholesky decomposition for $u_{\text{IC, BC}} := \left( u \mid \bm{\Lop_{\text{IC, BC}}}[u] + \bm{\varepsilon_{\text{IC, BC}}} = \bm{y_{\text{IC, BC}}} \right)$.
Then, we use $u_{\text{IC, BC}}$ as a new prior for which we use IterGP to integrate the PDE observations.
In other words, we apply IterGP with CG actions to compute $u_{\text{IBVP}} := \left( u_{\text{IC, BC}} \mid \bm{\Lop_{\text{PDE}}}[u] + \bm{\varepsilon_{\text{PDE}}} = \bm{y_{\text{PDE}}} \right)$.
\citet{Wenger2022} show that IterGP generalizes the Cholesky decomposition through unit vector actions.
Thus, in the sense of IterGP, this strategy constructs a deflation-based preconditioner which projects onto the subspace of the search space that satisfies the initial and boundary observations.
All further CG iterations are then isolated to the PDE and cannot violate initial or boundary observations.
This combines exact computations \textit{where they are affordable} with approximate computations \textit{where they are necessary}.
As a consequence, there is no computational uncertainty for the initial and boundary conditions.

\section{Related work} \label{sec:related-work}
\textbf{Physics-Informed Machine Learning}
Popular machine learning based approaches to PDE solving include physics-informed neural networks \citep{Raissi2019}, (Fourier) neural operators \citep{Li2020, Lu2021}, and neural ODEs \citep{Chen2018}.
The latter two rely on internal calls to low-level internal numerical methods, either to produce training data (for neural operators) or to solve ODEs (for NODEs).
Our method is instead directly related to classic stand-alone numerical PDE solvers.
This makes it most directly related to PINNs, but the analytic nature of our method allows for the significant computational savings outlined above.

\textbf{Probabilistic Numerics (PN)}
Our work fits into the framework of PN methods \citep{Hennig2015}, where several variations of GP-based PDE solvers have been considered.
\citet{Cockayne2017b} propose a solver based on collocation and also consider inverse problems.
A generalization of collocation to nonlinear PDEs is given by \citet{Chen2021} through a quadratic optimization problem with nonlinear constraints.
Other variants include multistep methods \citep{Raissi2017} or the PN method of lines \citep{Kramer2022}.

\textbf{MWRs}
\citet{Pfoertner2023} prove more generally that affine GP inference generalizes all methods of weighted residuals (MWRs), of which FVM is a special case.
We leverage this theoretical result to provide a practical framework for efficient inference on this observation class.
To the best of our knowledge, our work is the first to focus on practical aspects of FVM observations for GP inference.

\textbf{Scalability}
Other approaches towards scalable GP-based PDE solvers include approximations through Fourier features \citep{Mou2022}, sparse Cholesky decompositions \citep{Chen2023}, inducing points \citep{Meng2023}, and a stochastic proximal mini-batch approach \citep{Yang2023}.
The use of Kronecker structure has been proposed previously in the context of GPs \citep{Wilson2015}, PDE solvers \citep{Gavrilyuk2019}, and GP-based PDE solvers \citep{Kramer2022b, Fang2023}.
Our work complements this view, using a new observation class and an efficient iterative solver that quantifies its computational uncertainty.

\section{Experiments} \label{sec:experiments}
Here we investigate the utility and scaling behavior of our GP-FVM framework.
First, we run benchmarks to compare to collocation and deep learning-based approaches (\cref{sec:collocation-comparison}).
Afterwards, we test our methods on simulations on the scale of real-world physical processes (\cref{sec:experiments-tsunami}).

\textbf{Implementation}
An efficient implementation of GP-FVM is available as \href{https://github.com/timweiland/gp-fvm}{\underline{open-source Python code}}.
We build on source code from \citet{Pfoertner2023}, which in turn leverage the ProbNum package \citep{Wenger2021}.
The iterative solvers in particular are designed for execution on a GPU using PyTorch \citep{Paszke2019}.
The benchmarks in \cref{sec:collocation-comparison} were run on one Intel Xeon Gold 6240 CPU.
The tsunami simulation in \cref{sec:experiments-tsunami} was run on one NVIDIA A100 GPU.

\subsection{Comparative Benchmarks}
\label{sec:collocation-comparison}
\textbf{Setup}
We evaluate GP-FVM on three different IBVP classes.
From PDEBench \citep{Takamoto2022}, we use the \textbf{a)} 1D Advection and \textbf{b)} 2D Darcy Flow benchmarks, which are both linear problems parameterized by a scalar $\beta$.
\textbf{a)} is a spatiotemporal problem, whereas \textbf{b)} is only spatial and thus has no initial condition.
For fixed $\beta$, each benchmark has 10000 different IBVPs.
\textbf{a)} and \textbf{b)} are both 2D problems.
Hence, for further evaluation, we construct \textbf{c)} our own 3D benchmark based on the 2D wave equation, which has one temporal dimension and two spatial dimensions.
\crefrange{appendix:1d-advection}{appendix:2d-wave} contain more details as well as further experiments.

\textbf{Evaluation Metric}
For each IBVP, we compute the root mean squared error (RMSE) or the maximum absolute error (MAE) between the GP posterior mean and the true solution on domain-covering test points $\bm{X^{\mathrm{Test}}} \in \R^{N_{\mathrm{Test}} \times d}$.
We then report the mean of these metrics across all IBVPs.

\textbf{Models}
For each benchmark, we evaluate GP-based PDE solvers based on \textcolor{collocationblue}{collocation} and \textcolor{subdomainred}{GP-FVM}.
To assess the scaling behavior, we evaluate different amounts of observations $N_{\mathrm{PDE}} \in \N$.
For collocation, this is the number of collocation points; for GP-FVM, it is the number of finite volumes.
In both cases, we use discretization schemes that are uniformly distributed across the domain.
Details on this setup are contained in \crefrange{appendix:1d-advection}{appendix:2d-wave}.
Additionally, for \textbf{a)} we compare to deep learning-based models using the values reported in \citet{Takamoto2022}.

\begin{figure*}
  \centering
  \includegraphics{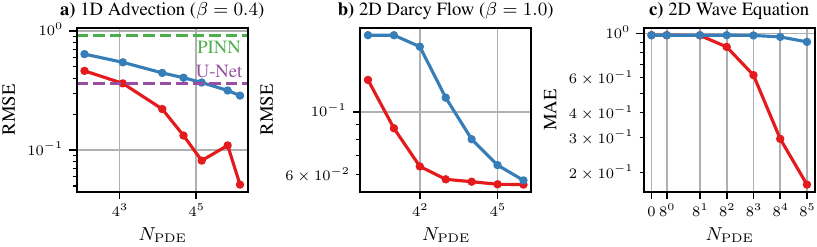}
  \caption{\textbf{Scaling behavior of \textcolor{subdomainred}{GP-FVM} compared to \textcolor{collocationblue}{collocation} for various problem classes.} The plots show the mean RMSE across all IBVPs in each problem class. For \textbf{a)}, we also compare to neural network baselines from PDEBench.}
  \label{fig:pdebench-benchmark}
\end{figure*}

\textbf{Results:}
\textbf{In all cases, \textcolor{subdomainred}{GP-FVM} outperforms \textcolor{collocationblue}{collocation}.}
\textcolor{subdomainred}{GP-FVM} generally requires several orders of magnitude fewer observations than \textcolor{collocationblue}{collocation} to reach the same solution quality.
The gap is particularly substantial for \textbf{a)} and \textbf{c)}.
For these benchmarks, the log-log plot implies different power-law relationships between \textcolor{subdomainred}{GP-FVM} and \textcolor{collocationblue}{collocation}.
\cref{fig:collocation-vs-fvm} provides some intuition on why this might be the case:
Where for \textcolor{collocationblue}{collocation} the uncertainty collapses in the derivative space, for \textcolor{subdomainred}{GP-FVM} it collapses directly in the original space.
The idea in general is that the integral yields observations in a space that is closer to the solution space than for \textcolor{collocationblue}{collocation}.
For \textbf{b)}, the difference vanishes for an increasing number of observations.
We attribute this to the highly discrete right-hand side function present for this benchmark, which may indeed require targeted point-wise observations to capture properly.

\textbf{Comparability}
Collocation can leverage Kronecker structure in the same way as GP-FVM.
Thus, for equal $N_{\mathrm{PDE}}$, both methods have comparable compute and equal memory requirements.

\subsection{Real-world example: Tsunami propagation} \label{sec:experiments-tsunami}
The results of \cref{sec:collocation-comparison} demonstrate the improved scaling behavior of GP-FVM over collocation.
In the following, we investigate to which extent GP-FVM is a feasible choice for less trivial real-world physical processes.
As an example, we choose a fluid dynamics problem.

\textbf{Setup}
We use the linearized 2D shallow-water equations (\crefrange{eq:sw1}{eq:sw3}, \cref{appendix:tsunami}) to model the propagation of a fictional tsunami towards a coast.
We propose a scenario where an earthquake causes an anisotropic Gaussian-shaped displacement of water (see \cref{fig:tsunami}).
An additional challenge of this setup is that \cref{eq:sw1} uses a variable coefficient $H(t, x, y)$ due to the seabed topography.
For details on these topics, refer to \cref{appendix:tsunami}.

The results (\cref{fig:tsunami}) demonstrate that GP-FVM yields a plausible posterior mean.
The arrival time of the synthetic tsunami is similar to that of a real tsunami observed in the past in this coastal area.

\textbf{Linear system scale}
The simulation is based on $918000$ FVM observations and an overall Gram matrix with $936672^2$ entries.
Based on the results from \cref{sec:collocation-comparison}, collocation would likely have required a prohibitively larger amount of observations to obtain a solution of this quality.
To solve this large problem, we first solve for the initial and boundary observations using a Cholesky decomposition (see \cref{sec:cholesky-preconditioner}).
Then we run IterGP with CG actions for a budget of $1250$ iterations.

\textbf{Uncertainty}
The posterior uncertainty (for which \cref{appendix:tsunami} contains dedicated plots) reflects the computational uncertainty induced by early termination of the solver.
We observe that the posterior covariance is underconfident in relation to the posterior mean quality.
This is a well-known issue with probabilistic linear solvers \citep{Cockayne2019b}.
Effectively, the computational uncertainty cannot match the fast convergence of CG.
To alleviate this issue, we perform $500$ additional iterations using actions targeted to uncertainty reduction around the shore, detailed in \cref{appendix:targeted-uncertainty-reduction}.
Additionally, we observe a collapse of uncertainty at $t = 0$ and at the boundaries. These observations are solved by a Cholesky decomposition, causing no computational uncertainty.

\begin{figure}
  \includegraphics{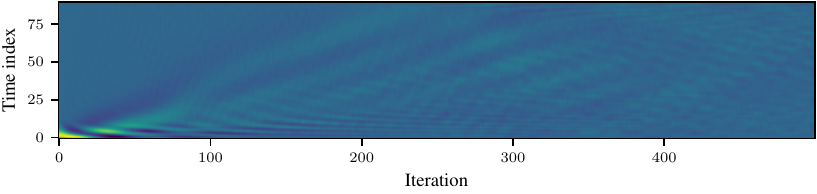}
  \caption{\textbf{IterGP actions show how the solver progresses through the time dimension.}
    Depicted are the first $500$ actions of IterGP with a CG policy for the tsunami example.
    Each action is reduced to the FVM components, and then summed up over the spatial dimensions.
    The result is a vector containing the weight by which each time slice is targeted.}
  \label{fig:shallow-action-matrix}
\end{figure}
\textbf{CG steps through time}
\cref{fig:shallow-action-matrix} depicts the behavior of CG for this problem.
CG weighs the subdomain observations by their current residual.
Initially, the residual is the largest for the volumes near $t = 0$.
Then, the solver gradually progresses through time.
The repeated diagonal lines are effectively caused by early observations requiring multiple iterations to sufficiently reduce the residual.
Meanwhile the progress from the previous iterations already propagates through time.

Including the post-iterations, the iterative linear solve requires about $15$ minutes, demonstrating the \textbf{feasibility of large-scale probabilistic PDE simulations} in our framework.

\section{Limitations}
\label{sec:limitations}
Like most classic PDE solvers, our approach requires iterative numerical linear algebra (in our case its probabilistic version, IterGP), which needs careful stabilization.
For the moment we rely on regular full reorthogonalization, which stabilizes the process, but incurs cost.
A second issue is that IterGP and other probabilistic linear solvers still tend to err towards overly conservative uncertainty estimates when using the purely exploratory CG policy.
\cref{appendix:targeted-uncertainty-reduction} suggests an approach to tackle this issue, but further treatment must remain beyond the scope of the present work.
On a lower level, Kronecker structure requires factorized discretization schemes.
While non-rectangular domains may be covered by such schemes, this may computational overhead over other discretization schemes.

\section{Conclusion}
We introduced a framework for efficient probabilistic inference on PDE solutions from volumetric (FVM) observations.
FVM observations scale favorably:
In our experiments, GP-FVM requires several orders of magnitude fewer observations than collocation to reach the same solution quality.
The matrix-free probabilistic numerical linear algebra techniques we proposed are highly efficient in solving the arising linear system.
Taken together, our setup scales probabilistic solutions to nontrivial simulation problems, providing a key step towards real-world applicability of probabilistic PDE simulators.
Such simulators remain important---also and especially in machine learning.
Work like ours explicitly describes simulation itself as an inference scheme, highlighting avenues for advancements
in scientific machine learning, including the design of deep emulators,
that would be difficult to realize if PDE simulation was left as a black box developed by a separate discipline.

\begin{ack}
The authors gratefully acknowledge co-funding by the European Union (ERC, ANUBIS, 101123955, Views and opinions expressed are however those of the authors only and do not necessarily reflect those of the European Union or the European Research Council. Neither the European Union nor the granting authority can be held responsible for them); the DFG Cluster of Excellence “Machine Learning - New Perspectives for Science”, EXC 2064/1, project number 390727645; the German Federal Ministry of Education and Research (BMBF) through the Tübingen AI Center (FKZ: 01IS18039A); the DFG SPP 2298 (Project HE 7114/5-1), and the Carl Zeiss Foundation, (project "Certification and Foundations of Safe Machine Learning Systems in Healthcare"), as well as funds from the Ministry of Science, Research and Arts of the State of Baden-Württemberg.
The authors thank the International Max Planck Research School for Intelligent Systems (IMPRS-IS) for supporting TW and MP.
\end{ack}

\bibliographystyle{plainnat}
\bibliography{main}

\newpage
\appendix

\section{Affine Gaussian Process Inference, in Detail} \label{appendix:affine-inference}
In the following, we elaborate on the assumptions required for the results in \cref{sec:affine-inference} to hold.
The following theory is based on Section 4 of \citet{Pfoertner2023}.

First, some assumptions on the prior are necessary.
This requires the following definition:
\begin{definition}
  A \textbf{reproducing kernel Banach space} (RKBS) is a Banach space of real-valued functions on a non-empty set $\X$ on which all point evaluation functionals are continuous.
\end{definition}
\begin{assumption} \label{assumption-rkbs-gaussian-randvar}
  $f \sim \mathcal{GP}(m, k)$ is a Gaussian process with index set $\X$ on $(\Omega, \mathcal{F}, P)$, mean function $m: \X \rightarrow \R$ and covariance function $k: \X \times \X \rightarrow \R$.
  The sample paths of $f$ lie in a separable RKBS $\mathbb{B}$ and $\omega \mapsto f(\cdot, \omega)$ is a $\mathbb{B}$-valued Gaussian random variable.
\end{assumption}
Under \cref{assumption-rkbs-gaussian-randvar}, we can condition a GP on observations obtained through linear operators.
The following theorem uses the notation from \crefrange{eq:Lk-def}{eq:LkL-def}:
\begin{theorem}\label{theorem:theone}
  Let $\bm{\Lop}: \mathbb{B} \rightarrow \Rn$ be a bounded linear operator.
  Under \cref{assumption-rkbs-gaussian-randvar}, the following results hold. \\
  Viewing $f$ as a $\mathbb{B}$-valued Gaussian random variable, the application of $\mathcal{L}$ yields
  \begin{equation}
    \bm{\Lop}[f] \sim \mathcal{N}(\bm{\Lop}[m], \bm{\Lop} k \bm{\Lop^{\prime}}).
  \end{equation}
  Let $\bm{\varepsilon} \sim \mathcal{N}(\bm{\mu}, \bm{\Sigma})$ be an $\Rn$-valued Gaussian random variable such that $\bm{\varepsilon} \perp f$, and let $\bm{y} \in \Rn$.
  Define $\bm{G} := \bm{\Lop} k \bm{\Lop^{\prime}} + \bm{\Sigma} \in \R^{n \times n}$. Then:
  \begin{equation}
    f \mid \bm{\Lop}[f] + \bm{\varepsilon} = \bm{y} \sim \GP(m^{*}, k^{*}),
  \end{equation}
  with
  \begin{align}
    m^{*}(\bm{x})                 & := m(\bm{x}) + \bm{\Lop} k(\bm{x})^{\tran} \bm{G}^{\pinv} (\bm{y} - (\bm{\Lop} [m] + \bm{\mu})),                        \\
    k(\bm{x_{1}}, \bm{x_{2}}) & := k(\bm{x_{1}}, \bm{x_{2}}) - \bm{\Lop} k(\bm{x_{1}})^{\tran} \bm{G}^{\pinv} \bm{\Lop} k(\bm{x_{2}}).
  \end{align}
\end{theorem}

\cref{theorem:theone} gives closed-form expressions for the posterior obtained by conditioning a GP on linear operator observations of its paths.
\citet{Pfoertner2023} also state the equivalent result for when the codomain of $\mathcal{L}$ is another real separable RKBS.
Classic GP regression on point evaluations then corresponds to the special case $\mathcal{L} = \mathrm{id}_{\mathbb{B}}$.

\section{Theoretical justification} \label{appendix:theoretical-justification}
For affine GP inference based on these ideas to be theoretically justified, the introduced priors must satisfy \cref{assumption-rkbs-gaussian-randvar} and the linear operators must be bounded on their sample spaces.

\citet{Pfoertner2023} show that if $f$ is a GP with $\mathrm{paths}(f) \subset C^{\bm{\beta}}(\bar{\mathbb{X}})$ $P$-almost surely ($\bm{\beta} \in \N_0^{d}$), then $f$ fulfills \cref{assumption-rkbs-gaussian-randvar}. As mentioned in \cref{sec:multi-dim}, this is the case for the tensor product covariance function.

For the multi-dimensional codomain case, our choice of covariance function amounts to modelling $d^{\prime}$ independent single-output GPs. Each of these single-output GPs has a sample space in $C^{\bm{\beta}}(\bar{\mathbb{X}})$ for some $\bm{\beta} \in \N_{0}^{d}$. As a consequence, the overall sample space $\mathbb{B}$ is isometrically isomorphic to a subset of the Cartesian product $\bigtimes_{i=1}^{d^{\prime}} C^{\beta_{i}}(\bar{\mathbb{X}})$, which is again a separable RKBS under an appropriate norm. Thus, $f$ is a Gaussian random variable whose values $P$-almost surely lie in $\mathbb{B}$. Hence, \cref{assumption-rkbs-gaussian-randvar} is fulfilled. For more details on the multi-dimensional codomain case, refer to \citet[Remark 4.2]{Pfoertner2023}.

We also need to prove that the linear operator we described is bounded on the sample space.
The linear operator is the composition of an integral over a finite box volume with a linear differential operator.
Because the sample space consists of functions in $C^{\bm{\beta}}(\bar{\mathbb{X}})$, we know that all partial derivatives appearing in the PDE are bounded and uniformly continuous (assuming $\bm{\beta}$ is sufficiently large).
It immediately follows that $\diffopD$ is bounded as a sum of bounded linear operators.

Furthermore, it is trivial to show that the integral of bounded and uniformly continuous functions over a finite box volume is bounded:
\begin{align}
  \left\lvert \int_{V} f(\mathbf{x}) d\mathbf{x} \right\rvert & \leq \int_{V} \lvert f(x) \rvert d\mathbf{x}                                                                                      \\
                                                              & \leq \int_{V} \lVert f \rVert_{C^{0}(\bar{\mathbf{X}})} d\mathbf{x}                                                               \\
                                                              & = \int_{a_{1}}^{b_{1}} \dots \int_{a_{d}}^{b_{d}} \lVert f \rVert_{C^{0}(\bar{\mathbf{X}})} dx_{d} \dots dx_{1} \\
                                                              & = \lVert f \rVert_{C^{0}(\bar{\mathbf{X}})} \prod_{i=1}^{d} (b_{i} - a_{i})                                                       \\
                                                              & \leq \lVert f \rVert_{C^{\bm{\beta}}(\bar{\mathbf{X}})} \prod_{i=1}^{d} (b_{i} - a_{i}).
\end{align}
Clearly, the integral is also linear. Hence, the operator that generates our observations is a bounded linear operator as a composition of bounded linear operators. This proves that our framework is valid for both the single-input and the multi-input cases.

\section{Experiment details}
\label{appendix:experiment-details}
\subsection{1D Advection}
\label{appendix:1d-advection}

\textbf{Setup}
The 1D advection equation is a spatiotemporal first-order linear PDE that models the transport of a quantity by some fluid flow.
We use the 1D advection problem class from PDEBench \citep{Takamoto2022}, which is parameterized by the constant advection speed $\beta \in \R$.
Its initial condition consists of a super-position of random sine wave.
PDEBench offers several values of $\beta$ and provides 10000 different IBVPs for each, which differ in the random initial condition.
In our work, we use $\beta = 0.4$.
For further details, refer to \citet{Takamoto2022}.

\textbf{Observation Layout}
For the initial condition, we downsample the provided resolution of $1024$ by a factor of $10$, which yields $\approx 100$ observations.
For each of the two boundary points, we use $50$ observations uniformly spaced across the time dimension.
Note that the periodic boundary can be expressed in the framework of affine GP inference through the linear functional
\begin{equation}
  \Lop_{\mathrm{periodic}, t}[u] = u(t, 0) - u(t, 1),
\end{equation}
where the spatial domain is $\mathbb{X} = \left[ 0, 1\right]$. Then $\Lop_{\mathrm{periodic, t}}[u] = 0$ is equivalent to $u(t, 0) = u(t, 1)$.
For each observation class, the PDE observations are uniformly spread across the domain.
Concretely, the collocation points are uniformly spaced, and the finite volumes all have equal area and disjointly span the entire domain.

\textbf{Prior}
We use a zero mean and a tensor product covariance function with a $\nicefrac{3}{2}$-Matérn covariance function per dimension.

\textbf{Hyperparameter Optimization}
To find the lengthscales of the covariance functions, we optimize for low MSE on $10$ random IBVPs in the dataset.
To this end, we use the default Bayesian optimization method implemented in the Optuna hyperparameter optimization framework \citep{Optuna2019}, which we run for 100 trials.
We do this separately between collocation and GP-FVM and for each value of $N_{\mathrm{PDE}}$.

\textbf{Compute}
The experiment was run on one Intel Xeon Gold 6240 CPU on an internal cluster.
A full run of the entire experiment, which produces \cref{fig:pdebench-benchmark} \textbf{a)}, requires about 24 hours of runtime, though we ran the subtasks for each value of $N_{\mathrm{PDE}}$ in parallel.
We estimate that our overall experimentation on this task during our research did not require more than about 28 hours of runtime in total.

\subsubsection{Additional results}
\begin{figure*}
  \centering
  \includegraphics{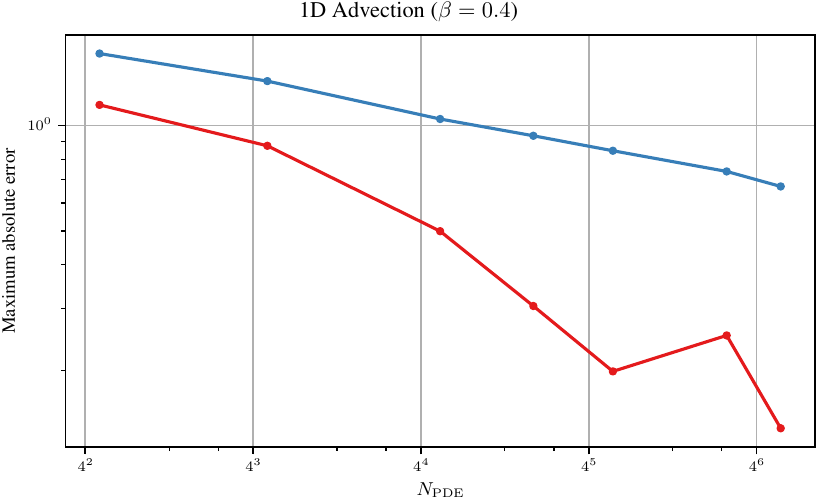}
  \caption{\textbf{Scaling behavior of \textcolor{subdomainred}{GP-FVM} compared to \textcolor{collocationblue}{collocation} for 1D Advection with $\beta = 0.4$.} The plot shows the mean maximum absolute error (MAE) across all IBVPs in each problem class.}
  \label{fig:advection-04-mae}
\end{figure*}
\begin{figure*}
  \centering
  \includegraphics{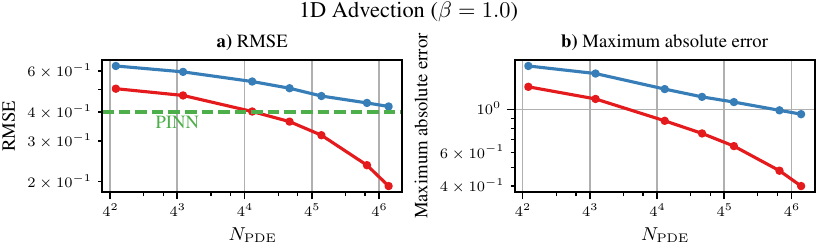}
  \caption{\textbf{Scaling behavior of \textcolor{subdomainred}{GP-FVM} compared to \textcolor{collocationblue}{collocation} for 1D Advection with $\beta = 1.0$.} The plots show \textbf{a)} the mean RMSE and \textbf{b)} the mean MAE across all IBVPs in each problem class. For \textbf{a)}, we also compare to the PINN baseline from PDEBench.}
  \label{fig:advection-10-rmse-mae}
\end{figure*}
\crefrange{fig:advection-04-mae}{fig:advection-10-rmse-mae} depict additional results for this problem class.
We observe a similar gap between \textcolor{subdomainred}{GP-FVM} and \textcolor{collocationblue}{collocation} both in terms of MAE for $\beta = 0.4$ as well as in terms of both RMSE and MAE for $\beta = 1.0$.
Note that for the MAE we cannot compare to the baselines from PDEBench as they form the mean over the time dimension rather than the maximum.
The U-Net baseline from PDEBench outperforms GP-FVM and collocation at an RMSE of $1.2 \times 10^{-2}$, which we do not plot in \cref{fig:advection-10-rmse-mae} for the sake of legibility.
Additionally, we note that for $\beta = 1.0$ we doubled the number of boundary conditions per boundary point from $50$ to $100$ to account for the increased speed of the PDE dynamics.

\clearpage

\subsection{2D Darcy Flow}
\label{appendix:2d-darcy}

\textbf{Setup}
2D Darcy Flow is a spatial second-order linear PDE that describes fluid flow through a porous medium.
We use the 2D Darcy Flow problem class from PDEBench \citep{Takamoto2022}, which is parameterized by a constant force term $\beta \in \R$.
Since there is no time dimension, there is also no initial condition.
Instead, the PDE is directly influenced by the variable viscosity term.
This term is given by a discontinuous function over the 2D space which differs for each of the 10000 IBVPs per $\beta$.
In our work, we use $\beta = 1.0$.
The Dirichlet boundary condition is given by $u(x) = 0$ $(x \in \partial \mathbb{X})$.
For further details, refer to \citet{Takamoto2022}.

\textbf{Difference to deep learning-based solvers}
An important difference to the deep learning-based solvers used in the work of \citet{Takamoto2022} is that we directly solve for the steady state.
By contrast, the deep learning-based solvers start with a random field initial condition and then solve a temporal evolution equation until reaching a steady state.
Hence, we do not compare to their metrics.

\textbf{Observation Layout}
For the boundary condition, we use $30$ uniformly spaced observations per edge of the domain.
For each observation class, the PDE observations are uniformly spread across the domain.
Concretely, the collocation points are uniformly spaced, and the finite volumes all have equal area and disjointly span the entire domain.

\textbf{Variable parameter}
This benchmark uses a PDE with a variable parameter.
For collocation, this does not change anything.
For GP-FVM, we approximate the integral in the same way as described for the tsunami experiment (\cref{appendix:tsunami}), which also has a variable parameter.

\textbf{Prior}
We use a zero mean and a tensor product covariance function with a $\nicefrac{5}{2}$-Matérn covariance function per dimension.

\textbf{Hyperparameter Optimization}
To find the lengthscales of the covariance functions, we optimize for low MSE on $10$ random IBVPs in the dataset.
To this end, we use the default Bayesian optimization method implemented in the Optuna hyperparameter optimization framework \citep{Optuna2019}, which we run for 30 trials.
We do this separately between collocation and GP-FVM and for each value of $N_{\mathrm{PDE}}$.
To simplify the optimization problem, we use the same lengthscale for the $x$ and $y$ dimension.

\textbf{Compute}
The experiment was run on one Intel Xeon Gold 6240 CPU on an internal cluster.
A full run of the entire experiment, which produces \cref{fig:pdebench-benchmark} \textbf{b)}, requires about 23 hours of runtime, though we ran the subtasks for each value of $N_{\mathrm{PDE}}$ in parallel.
We estimate that our overall experimentation on this task during our research did not require more than about 29 hours of runtime in total.

\subsubsection{Additional results}
\begin{figure*}
  \centering
  \includegraphics{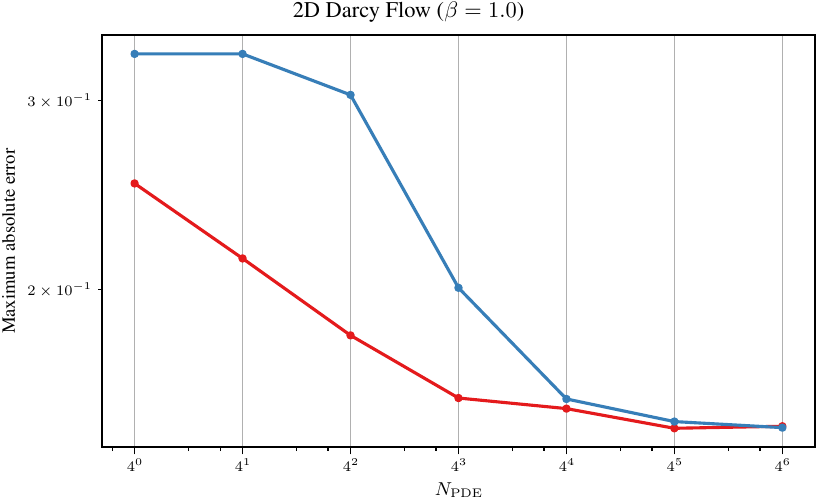}
  \caption{\textbf{Scaling behavior of \textcolor{subdomainred}{GP-FVM} compared to \textcolor{collocationblue}{collocation} for 2D Darcy Flow with $\beta = 1.0$.} The plot shows the mean maximum absolute error (MAE) across all IBVPs in each problem class.}
  \label{fig:darcy-10-mae}
\end{figure*}
\begin{figure*}
  \centering
  \includegraphics{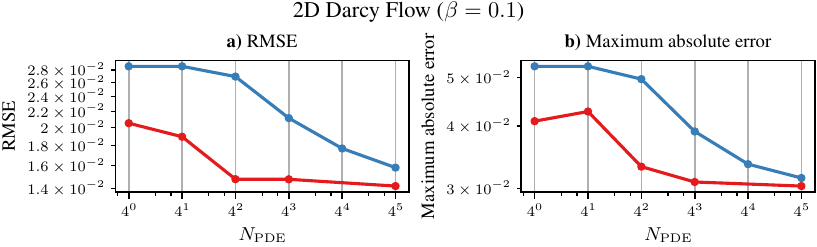}
  \caption{\textbf{Scaling behavior of \textcolor{subdomainred}{GP-FVM} compared to \textcolor{collocationblue}{collocation} for 2D Darcy Flow with $\beta = 0.1$.} The plots show \textbf{a)} the mean RMSE and \textbf{b)} the mean MAE across all IBVPs in each problem class.}
  \label{fig:darcy-01-rmse-mae}
\end{figure*}
\begin{figure*}
  \centering
  \includegraphics{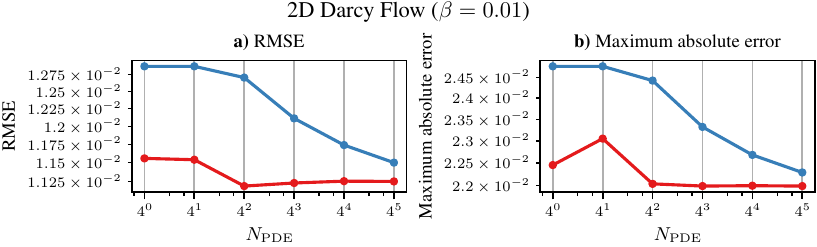}
  \caption{\textbf{Scaling behavior of \textcolor{subdomainred}{GP-FVM} compared to \textcolor{collocationblue}{collocation} for 2D Darcy Flow with $\beta = 0.01$.} The plots show \textbf{a)} the mean RMSE and \textbf{b)} the mean MAE across all IBVPs in each problem class.}
  \label{fig:darcy-001-rmse-mae}
\end{figure*}
\crefrange{fig:darcy-10-mae}{fig:darcy-001-rmse-mae} depict additional results for this problem class.
We observe similar behavior as described for \cref{fig:collocation-vs-fvm} \textbf{b)}: There is a large initial gap between \textcolor{subdomainred}{GP-FVM} and \textcolor{collocationblue}{collocation} initially, which diminishes as the number of observations grows.
Notably, the MAE increases for GP-FVM for $\beta = 0.1$ and $\beta = 0.01$.
We assume that this is another effect of the highly discrete right-hand side function present in this problem class.

\clearpage

\subsection{2D Wave Equation Benchmark}
\label{appendix:2d-wave}
The 2D wave equation is given by:
\begin{equation}
  \frac{\partial^2}{\partial t^2} u(t, \bm{x}) = c^{2} \Delta u(t, \bm{x}), \label{eq:wave}
\end{equation}
where $c \in \R$ is the wave speed.
We choose a wave speed of $c = 1$.
We consider functions $u(t, x, y)$ on the domain $\domain = [0, 2] \times [0, 1] \times [0, 1]$.

\textbf{Initial condition}
We consider a problem class with initial conditions
\begin{gather}
  u(0, x, y) = \sum_{i=1}^{2} \sum_{i=1}^{2} C_{ij} \cdot \sin \left( i  \pi  x \right) \cdot \sin \left( j  \pi  y \right), \quad (x, y \in [0, 1]), \label{eq:wave-initial-condition} \\
  \frac{\partial}{\partial t} u(0, x, y) = 0, \quad (x, y \in [0, 1])
\end{gather}
for
\begin{equation}
  \label{eq:2d-wave-coefficients}
  \mathbf{C} = \begin{pmatrix} \tilde{c}_{1} & \tilde{c}_{2} \\ \tilde{c}_{3} & \tilde{c}_{4} \end{pmatrix}, \quad \mathbf{\tilde{c}} \sim \mathcal{N} \left(\begin{pmatrix} 1 & 0.5 & 0.5 & 0 \end{pmatrix}^{\tran} \!, \, \text{diag} \left(\begin{pmatrix} 0.1^{2} & 0.2^{2} & 0.2^{2} & 0.3^{2} \end{pmatrix}^{\tran} \right) \right).
\end{equation}

\textbf{Boundary condition}
We use Dirichlet boundary conditions
\begin{equation}
  \label{eq:wave-dirichlet}
  u(t, x, y) = 0, \quad (x, y \in \partial \domain)
\end{equation}
where $\partial \domain$ denotes the boundary of $\domain$.

This yields a well-posed hyperbolic IBVP, for which the unique solution can be obtained via the method of separation of variables.

\textbf{Observation Layout}
For the initial condition, we use $8$ uniformly spaced observations per dimension for a total of $64$ observations.
For each edge of the boundary, we use $20$ uniformly spaced observations along the temporal dimension and $10$ uniformly spaced observations along the spatial dimension.
For each observation class, the PDE observations are uniformly spread across the domain.
Concretely, the collocation points are uniformly spaced, and the finite volumes all have equal area and disjointly span the entire domain.

\textbf{Prior}
We use a zero mean and a tensor product covariance function with a $\nicefrac{5}{2}$-Matérn covariance function per dimension.

\textbf{Hyperparameter Optimization}
For each observation class and IBVP, we cycle through different lengthscale combinations and choose the one that yields the lowest error.
For the temporal covariance function, we cycle through $5$ uniformly spaced lengthscales in the interval $[0.1, 2.0]$ and for the spatial covariance functions, we cycle through $5$ uniformly spaced lengthscales in the interval $[0.1, 1.0]$. To save computational resources, we use the same lengthscale for the $x$ and $y$ dimensions.

\textbf{Compute}
Most of the experiment was run on one Intel Xeon Gold 6240 CPU on an internal cluster.
For the largest discretization size ($N_{\mathrm{PDE}} = 8^5$ in the figure), we use IterGP with CG actions on one NVIDIA GeForce RTX 2080 Ti GPU.
A full run of this entire experiment, which combines collocation and GP-FVM, performs hyperparameter optimization, and steps through all values of $N_{\mathrm{PDE}}$, requires about 18 hours.
We estimate that we spent about three such runs during our research on fixing bugs in the code until we converged at the results in the paper.

\subsection{Tsunami propagation}
\label{appendix:tsunami}

\textbf{PDEs}
We model the propagation of a tsunami using the linearized 2D shallow-water equations \citep{Vallis2017}:
\begin{align}
  \partialt h + H \left( \partialx u + \partialy v \right) & = 0, \label{eq:sw1} \\
  \partialt u - fv + g \partialx h + ku                    & = 0, \label{eq:sw2} \\
  \partialt v + fu + g \partialy h + kv                    & = 0. \label{eq:sw3}
\end{align}
The vertical deviation of a body of water from its mean height $H(t, x, y)$ is modelled by $h(t, x, y)$.
The water velocities in the $x$- and $y$-directions are modeled by $u(t, x, y)$ and $v(t, x, y)$, respectively.
The scalar parameters are the gravitational acceleration $g$, the Coriolis coefficient $f$ and the viscous drag coefficient $k$.

\textbf{Boundary condition}
At the boundary, we use radiation boundary conditions with wave speed $c(x, y) = \sqrt{g H(x, y)}$. For example, at $x = 0$, we use the condition
\begin{equation}
  \frac{\partial h}{\partial t} = c(0, y) \frac{\partial h}{\partial x},
\end{equation}
and similarly for the other boundaries. This condition effectively allows outward flow and rejects inward flow.

\textbf{Bathymetry}
For the the seabed topography (also known as bathymetry), we use data from the GEBCO 2023 grid \citep{GEBCO2023} of a rectangular crop of Japan's coastal area.
The data is included in the supplementary material.

\textbf{Finite volume layout}
We use finite volumes of equal volume that disjointly span the entire domain.
Each volume has approximately 1km length along the $x$ axis, 5km length along the $y$ axis and 20 seconds span across the temporal dimension.
This results in $90 \cdot 170 \cdot 20 = 306000$ volumes, and each volume yields three observations for the three PDEs.
We chose this imbalance between the $x$ and $y$ axes because the initial condition exhibits higher frequencies along the $x$ axis.

\textbf{IBVP scale}
We consider a simulation length of 30 minutes. For the initial condition, we use uniformly spaced observations with a spacing of 3km per spatial dimension. We also use uniformly spaced observations for the boundary conditions with a spatial spacing of 5km and a temporal spacing of 30 seconds.

\textbf{Prior}
We use an independent multi-output covariance function consisting of three tensor products of $\nicefrac{5}{2}$-Matérn covariance functions with lengthscales 60, 6 and 10 for the $t$, $x$ and $y$ dimensions.

\textbf{Variable bathymetry}
\cref{eq:sw1} uses a variable coefficient $H(t, x, y)$.
It is constant with respect to time, but not with respect to space due to the variable bathymetry.
To handle this challenge, we assume that the bathymetry is piecewise constant within each subdomain used to discretize the PDE.
This amounts to forming Hadamard products of the Kronecker products in \cref{eq:kronecker-structure} with rank-one matrices formed by the seabed topography.
At the resolution used for the subdomain observations, the piecewise constant approximation is nearly exact for the $x$ dimension and not too inaccurate for the $y$ dimension.

\begin{figure}
  \centering
  \includegraphics{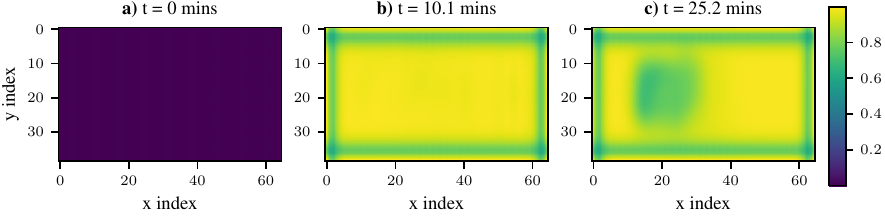}
  \caption{\textbf{Posterior covariance of the tsunami simulation.} The heatmaps depict the marginal posterior covariance at the same discrete locations used to create \cref{fig:tsunami}.}
  \label{fig:tsunami-cov}
\end{figure}
\textbf{Uncertainty}
\cref{fig:tsunami-cov} shows the posterior marginal uncertainty from \cref{fig:tsunami}.
The uncertainty reduction in a rectangular area between \cref{fig:tsunami}b) and \cref{fig:tsunami}c) is caused by post-iterations using the policy described in \cref{appendix:targeted-uncertainty-reduction}. This policy is based on $10 \cdot 5 \cdot 5$ uniformly spaced points in the subdomain $[1200, 1800] \times [40, 75] \times [33.3, 66.6]$.

\textbf{Compute}
The experiment was run on one NVIDIA A100 GPU on an internal cluster.
One full run of the experiment requires about 1 hour of runtime, which includes data loading and writing as well as a high-resolution evaluation of the posterior.
The actual iterative linear solve requires about 15 minutes.
We estimate that we spent about 15 runs in total to fix bugs and tweak parameters during our research until we converged at the run presented in the paper.

\subsection{Dataset uncertainty}
\label{sec:dataset-uncertainty}
\begin{figure}
  \includegraphics{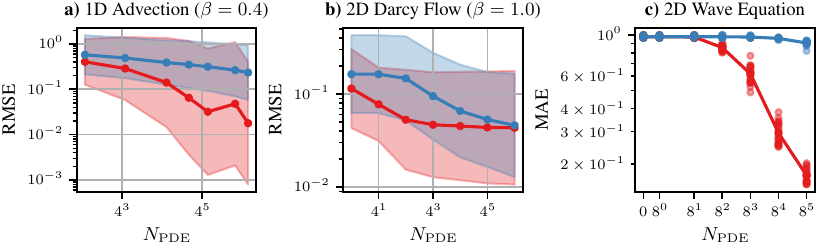}
  \caption{\textbf{Uncertainty in the benchmark datasets.}
    For \textbf{a)} and \textbf{b)}, we plot the $1.96 \cdot \sigma$ error bands.
  For \textbf{c)}, we plot all individual samples considered in the benchmark through transparent dots.}
  \label{fig:error-bars-fig}
\end{figure}
\cref{fig:collocation-vs-fvm} is based on fixed datasets where each IBVP is the result of a random initial condition.
\cref{fig:error-bars-fig} plots the resulting variance in the error metrics across the IBVPs.
For \textbf{a)} and \textbf{b)}, the means and standard deviations were computed in logarithmic space and then transformed back for visualization.
The reason is that the Normality of errors is a more reasonable assumption in logarithmic space for our results.
This is also one of the reasons why we do not include error bands in \cref{fig:collocation-vs-fvm}: Aside from hindering legibility, this transformation would prevent comparability to the metrics from PDEBench since the logarithm of a mean is generally not equal to the mean of a logarithm.

\section{Targeted uncertainty reduction}
\label{appendix:targeted-uncertainty-reduction}
As described in \cref{sec:experiments-tsunami}, the posterior uncertainty visualized in \cref{fig:tsunami} and \cref{fig:tsunami-cov} is underconfident.
In this section, we derive a simple technique to obtain better uncertainty quantification locally in targeted areas of the domain.

Consider \cref{eq:lin-cond-cov}. Computing the marginal posterior covariance exactly at $N^{*}$ points given by $\bm{X^{*}} \in \R^{N^{*} \times d}$ would require $N^{*}$ linear system solves, encapsulated by $\bm{G}^{\pinv} \underbrace{\bm{\Lop} k(\bm{X^{*}})}_{\in \R^{N \times N^{*}}}$.

Instead, we focus on a subsample $\bm{\tilde{X}} \in \R^{\tilde{N} \times d}$ with $\tilde{N} \ll N^{*}$. In the tsunami example, we chose uniformly spaced points towards the end of the simulation, close to the shore. Even for a subsample we cannot afford $\tilde{N}$ accurate linear system solves. Instead, we leverage similarities between related linear systems by sampling linear combinations of the columns of $\bm{\Lop} k(\bm{\tilde{X}}) \in \R^{N \times \tilde{N}}$.

Concretely, we sample a weight vector $\bm{\beta} \sim \text{Dir}(1, \dots, 1)$ from a Dirichlet distribution and consider the linear combination $\bm{\Lop} k(\bm{\tilde{X}}) \bm{\beta} \in \R^N$. IterGP iteratively builds up an inverse approximation, which we denote by $\bm{C_i}$ at iteration $i$. Then the residual with respect to the inverse approximation\footnote{Refer to \citet{Wenger2022} for details.} is given by
\begin{equation}
  \bm{r}_i := \bm{\Lop} k(\bm{\tilde{x}}) \bm{\beta} - \bm{G} \bm{C_i} \bm{\Lop} k(\bm{\tilde{x}}) \bm{\beta}.
\end{equation}
Using these residual vectors as IterGP actions implements CG with respect to the right-hand side $\bm{\Lop} k(\bm{\tilde{X}}) \bm{\beta}$. Thus, we iteratively solve $\bm{G}^{\pinv} \bm{\Lop} k(\bm{\tilde{X}}) \bm{\beta}$. If we sample a different $\bm{\beta}$ in each iteration, we effectively spread the solve across the entire area covered by $\bm{\tilde{X}}$.

The idea of using a Lanczos process on the right-hand side columns in \cref{eq:lin-cond-cov} to build up a low-rank approximation effective at uncertainty quantification has been considered previously by \citet{Pleiss2018}, which directly inspired our approach described here.
We start the Lanczos process indirectly through CG, and the low-rank approximation is automatically constructed by IterGP's inverse approximation.

\begin{figure}
  \centering
  \begin{tikzpicture}[y={(-1cm,0.5cm)},x={(1cm,0.5cm)}, z={(0cm,1cm)}]
    \coordinate (O) at (0, 0, 0);

    \foreach \x in {0,0.5,...,2}
    \foreach \y in {0,0.5,...,2}
      {
        \draw[color1] (\x,0) -- (\x,2);
        \draw[color1] (0,\y) -- (2,\y);
      }

    \foreach \x in {0,1,...,2}
    \foreach \y in {0,1,...,2}
      {
        \draw[color2] (\x,0,1.5) -- (\x,2,1.5);
        \draw[color2] (0,\y,1.5) -- (2,\y,1.5);
      }

    \foreach \x in {0,2}
    \foreach \y in {0,2}
      {
        \draw[color3] (\x,0,3) -- (\x,2,3);
        \draw[color3] (0,\y,3) -- (2,\y,3);
      }

    \draw[dotted, line width=0.5mm] (1,1,0) -- (1,1,1.5);
    \draw[dotted, line width=0.5mm] (1,1,1.5) -- (1,1,3);
    \node at (3, 1.5, 2.5) {$\textcolor{color3}{\mathcal{V}_0}$};
    \node at (3, 1.5, 1.0) {$\textcolor{color2}{\mathcal{V}_1}$};
    \node at (3, 1.5, -0.5) {$\textcolor{color1}{\mathcal{V}_2}$};
  \end{tikzpicture}
  \caption{\textbf{FVM discretization schemes nested through refinement.} Solutions that satisfy $\textcolor{color1}{\mathcal{V}_2}$ must also satisfy $\textcolor{color2}{\mathcal{V}_1}$ and $\textcolor{color3}{\mathcal{V}_0}$.}
  \label{fig:multigrid-pyramid}
\end{figure}
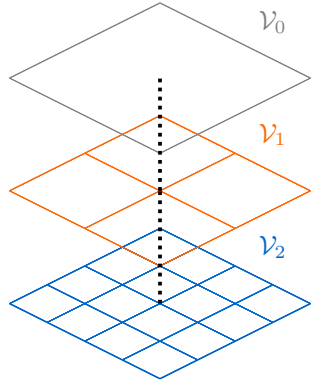
\section{Multigrid preconditioners} \label{sec:multigrid-preconditioner}
Multigrid methods are among the fastest classic PDE solvers that exist \citep{Trottenberg2000}.
They use smoothing properties of certain iterative methods and cycle between different discretization resolutions to accelerate convergence.

An analogue in our framework may be derived from redundancies in the FVM observations.
Consider a volume $V \subset \domain$ and a decomposition $V = \dot{\bigcup}_{j=1}^{k} \tilde{V}_{j}$ into disjoint smaller volumes $\tilde{V}_{j}$ ($j \in \{1, \dots, k \}$, $k \in \N$). Then for any integrable function $f: V \rightarrow \R$, we have
\begin{equation}
  \label{eq:subdomain-partition}
  \int_{V} f(x) dx = \sum_{i=1}^{k} \int_{\tilde{V}_{j}} f(x) dx.
\end{equation}

Consider nested FVM discretization schemes as in \cref{fig:multigrid-pyramid}.
The coarse schemes are partitioned by the finer schemes.
Thus, due to \cref{eq:subdomain-partition}, any solution that satisfies a fine scheme must also satisfy the coarser schemes.
Conditioning on sufficiently coarse discretization schemes can be performed exactly with a Cholesky decomposition.
Thus, we propose to further condition $u_{\text{IC, BC}}$ (\cref{sec:cholesky-preconditioner}) on coarse FVM observations.
Afterwards, we may use IterGP with CG actions for the fine FVM observations as before.
The idea is that warmstarting CG with exact solves at a coarser resolution (which is a form of preconditioner) may further accelerate its convergence.

\section{Conservation Laws} \label{appendix:conservation-laws}
PDEs are often derived from conservation laws in integral form.
As an example, consider a fluid with density $\rho(t, \bm{x})$ and velocity $u(t, \bm{x})$.
Then conservation of mass with respect to a fixed control volume $V$ with surface $A$ is expressed by the equality \citep{Kundu2015}:
\begin{equation}
  \label{eq:mass-conservation}
  \frac{d}{dt} \int_V \rho(t, \bm{x}) dV = - \int_A \rho(t, \bm{x}) u(t, \bm{x}) \bm{n} dA.
\end{equation}
Intuitively, this means that the amount of mass contained inside a static volume may only change through inflow or outflow through its boundaries.

In the finite volume method, the domain is divided into a fixed number of control volumes.
Then, conservation laws are solved by balancing fluxes between neighboring volumes.
As a result, the finite volume method is inherently conservative.

Similarly, our method also divides the domain into volumes.
The difference is that our method is based on fulfillment of some PDE, on average, across the volume.
This PDE may be the differential form of a conservation law.
But there is no explicit balancing of fluxes, and thus, our method is not inherently conservative.

Nevertheless, there are situations where our method can directly express conservation laws.
Assume we model $\rho(t, \bm{x})$ by a GP as described in \crefrange{sec:one-dim}{sec:multi-dim}.
Then the left-hand side of \cref{eq:mass-conservation} can be computed in closed-form.
Due to our use of box volumes, the surface integral in \cref{eq:mass-conservation} is a one-dimensional integral.
For fluids with constant velocity at the surface, the right-hand side is thus also tractable.
A special case is a volume with closed boundaries that allow no outward flow, i.e. $u(t, \bm{x}) = 0$ at the boundaries.
In these situations, the fulfillment of a conservation law at discrete points in time may be expressed by subdomain observations.

In the spirit of probabilistic numerics, a conservation law can thus be considered as yet another type of information about the function we want to infer.
It may be used, for example, to construct priors with samples that are all (approximately) globally conservative of some quantity.
This prior may then be conditioned on PDE information to obtain a simulation in the posterior that respects the conservation law.

Even if the PDE implicitly already contains the corresponding conservation law this may be useful.
If the resolution of the PDE observations is too coarse, numerical dissipation \citep{Trefethen1996} may deteriorate the quality of the solution.
Explicitly ``injecting'' the conservation law into the prior may counteract numerical dissipation.

\end{document}